# Using Federated Machine Learning in Predictive Maintenance of Jet Engines

Asaph Matheus Barbosa[1*], Thao Vy Nhat Ngo[1†], Elaheh Jafarigol[1†], Theodore B. Trafalis[1†], Emuobosa P. Ojoboh[1†]

[1]Data Science, The University of Oklahoma, 660 Parrington Oval, Norman, 73019, OK, United States of America.

*Corresponding author(s). E-mail(s): matheus@ou.edu;
Contributing authors: thao.vy.n.ngo@ou.edu; elaheh.jafarigol@ou.edu; ttrafalis@ou.edu; emuobosa.p.ojoboh-1@ou.edu;
†These authors contributed equally to this work.

**Abstract**

The goal of this paper is to predict the Remaining Useful Life (RUL) of turbine jet engines using a federated machine learning framework. Federated Learning enables multiple edge devices/nodes or servers to collaboratively train a shared model without sharing sensitive data, thus preserving data privacy and security. By implementing a nonlinear model, the system aims to capture complex relationships and patterns in the engine data to enhance the accuracy of RUL predictions. This approach leverages decentralized computation, allowing models to be trained locally at each device before aggregating the learned weights at a central server. By predicting the RUL of jet engines accurately, maintenance schedules can be optimized, downtime reduced, and operational efficiency improved, ultimately leading to cost savings and enhanced performance in the aviation industry. Computational results are provided by using the C-MAPSS dataset which is publicly available on the NASA website and is a valuable resource for studying and analyzing engine degradation behaviors in various operational scenarios.

## 1 Introduction

Integrating privacy-preserving machine learning (ML) techniques—particularly Federated Learning (FL)—into Predictive Maintenance (PM), represents a significant shift in how industrial operations manage maintenance and data security. This approach has broad implications for operational efficiency, data privacy, regulatory compliance, and technological innovation. In FL, data are processed locally at the device or server level, drastically reducing the risk that sensitive information is exposed during transmission or in a centralized database (McMahan et al., 2017). This is crucial for industries where operational data may include proprietary or sensitive business information. By minimizing data centralization, federated learning decreases the vulnerability of systems to massive data breaches, a growing concern with the increasing incidences of cyber-attacks. This work investigates the application of FL to PM tasks, focusing on the utilization of Long Short-Term Memory (LSTM) networks for predicting engine faults. Our research is motivated by the increasing demand for efficient PM methodologies that can minimize downtime and reduce operational costs in various industries(Diamoutene et al., 2018). Privacy-preserving ML enables real-time data analysis directly on the machines where data is generated. This allows for immediate identification of potential issues, facilitating quicker responses to prevent failures. With the ability to analyze data across a network of devices without compromising privacy, organiza-tions can optimize maintenance schedules based on predictive insights, rather than reactive or scheduled maintenance strategies. This not only extends the life of equipment but also reduces unnecessary downtime. FL aligns with global data protection regulations towards treating privacy as a fundamental human right and establishing robust privacy protection mechanisms in the era of artificial intelligence (Tene and Polonetsky, 2011; Brandeis and Warren, 1890). The latest update of the National Artificial Intelligence R&D Strategic Plan by the White House in 2023 underscores the significance of FL in addressing data privacy and security concerns. [1] This plan elaborates on long-term investment strategies in responsible AI research, emphasizing the need for advancements in privacypreserving data sharing and the ongoing challenges within FL. The General Data Protection Regulation (GDPR) [2] in Europe, also emphasizes data minimization, privacy by design, and the principle of processing data close to its source. Since FL does not require data to leave its source, it simplifies compliance with laws that restrict cross-border data transfers, making it an attractive option for multinational corporations. In FL only essential model-related information is transmitted. Therefore, FL can significantly reduce the costs associated with data transmission. In addition, by processing data locally and not requiring a central repository for vast amounts of raw data, companies can save on storage costs.

Through iterative experimentation and parameter tuning, we refine our models' accuracy, utilizing FL to distribute the computational load and enhance data privacy. The results section provides an in-depth analysis of the models' performance. Federated learning allows for the development of highly tailored models that learn from diverse data sources without compromising sensitive information. This capability can drive innovation in PM technologies. Different entities, even competitors, can collaborate to improve predictive models without sharing sensitive data, accelerating industry-wide advancements in maintenance strategies. Managing FL across many devices and locations introduces complexity, especially when coordinating updates and maintaining consistent model performance across diverse environments. Implementing an FL system requires sophisticated infrastructure and a shift in traditional data management strategies, which might be challenging for some organizations. This study encompasses a process validation section, where we underscore our collaboration with industry experts to ensure that our research objectives align with practical applications and that our findings remain relevant. The use of privacy-preserving ML and FL in PM strengthens data security and enhances operational efficiencies and compliance with regulatory norms. These technologies are setting new standards for how industries approach maintenance tasks while safeguarding critical data. As adoption grows, they could redefine best practices for asset management across various sectors, promising a future where PM is both more effective and inherently secure. In summary, this study delves into the application of federated learning for PM, presenting a structured approach to model development employing LSTM networks. Our findings aim to contribute to the ongoing discussion on the potential of FL in industrial applications, particularly in enhancing PM strategies to achieve operational efficiency and reliability.

---

[1] National Artificial Intelligence Research and Development Strategic Plan: https://www.nitrd.gov/national-artificialintelligence-research-and-development-strategicplan-2023-update

[2] 2General Data Protection Regulation: https://gdpr-info.eu

This paper is organized as follows. In section 2 we discuss the background and definition of the PM problem. Section 3 explores the data used in our study. In section 4 we discuss the methodology of our approach. Section 5 provides experiment results and discussion. Finally section 6 gives the conclusion of the paper.

## 2 Definition and background of the problem

Aircraft maintenance historically has two main philosophies: reactive and proactive. Both are widely used due to their different advantages and disadvantages. Reactive maintenance (Stanton et al., 2023) describes the process of waiting for the life cycle of a part of an airplane subsystem to completely run out before repairing or replacing the faulty components. Proactive maintenance (Meissner et al., 2021) describes the process of scheduling regular maintenance, to repair/replace components before they become faulty. The advantage of reactive maintenance is that we can get 100% usage out of our parts, but the obvious disadvantage is that there is a high chance of component failure happening during flights. This can work for something noncritical like overhead cockpit lights, which would not force a flight to be grounded if they failed in flight. On the other hand, something like a High-Pressure Compressor (HPC) failure in flight could prove disastrous. In these scenarios, it is better to perform proactive maintenance, where the disadvantage is that we lose some usage from our components, but we limit the number of in-flight failures. In more recent years with the advancements of ML (Jiang et al., 2023), PM has become more popular as a third approach where we can use machine learning to schedule our maintenance for high-risk systems and still get close to 100% usage with a low-error model (Asif et al., 2022). There are two main problems with this approach: small fleets with small sample sizes to train their PM models and large fleets unwilling to share their plentiful data with competitors due to privacy concerns. The work done in this study applies FL to address both of these problems with a single solution.

## 3 Data

NASA Ames Prognostics Center of Excellence (PCoE) researchers conducted engine degradation simulations using the Commercial Modular Aero-Propulsion System Simulation (C-MAPSS) (Saxena et al., 2008). The C-MAPSS dataset is publicly available and readily accessible on the NASA website, which can serve as a valuable resource for studying and analyzing engine degradation behaviors in various operational scenarios. The data was converted to .csv and is stored on our GitHub repository.

The C-MAPSS dataset is an operational behavior dataset from different engines. It offers a detailed look into the normal operational conditions of engines, including the presence of noise. Each data entry in the dataset contains 26 columns, encompassing information such as unit number, time cycles, three operational settings, and 21 sensor measurements (see Appendix A for details). These data snapshots, taken during individual operational cycles, provide valuable insights into the engine's behavior. Table 1 provides detailed information about each node. Sensor measurements are observed to be contaminated with noise (Botre et al., 2019), which can potentially introduce inaccuracies and inconsistencies in the data analysis process.

The dataset is structured into four training and four testing datasets, each with varying numbers of trajectories, conditions, and fault modes. The training sets are designed to showcase examples of faults that grow in magnitude until system failure occurs, providing valuable learning opportunities for predictive maintenance and fault detection. The testing sets may end before system failure, allowing for the evaluation of predictive models under different scenarios.

**Table 1**: Trajectories and Conditions for each Node

| Agent Name | Train Size | Test Size | Sim. Cond. | Faults |
|---|---|---|---|---|
| FD001 | 100 | 100 | 1 | HPC |
| FD002 | 260 | 259 | 6 | HPC |
| FD003 | 100 | 100 | 1 | Fan/HPC |
| FD004 | 248 | 249 | 6 | Fan/HPC |

### 3.1 Heatmaps

Heatmaps are powerful visualization tools used to identify correlations between variables in a dataset, making them particularly useful for analyzing relationships among sensor measurements (Ebrahimi et al., 2024). By plotting a heat map of the correlation matrix, patterns of correlation (both positive and negative) between pairs of variables can be easily visualized through color gradients. In the plot below, the lightest (white/tan) and darkest (black) colors indicate the highest linear correlation among variables. The orange/red color indicates there is little to no linear correlation between the variables.

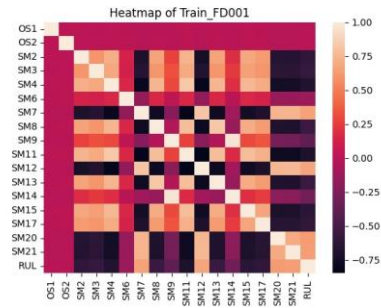

(a)

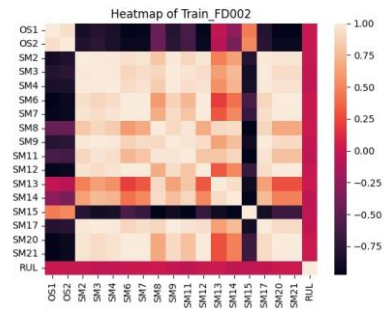

(b)

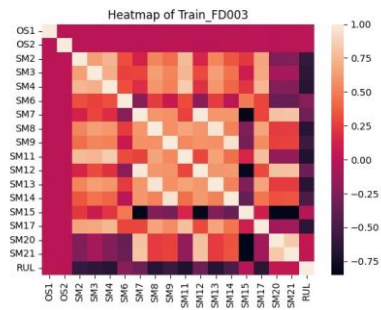

(c)

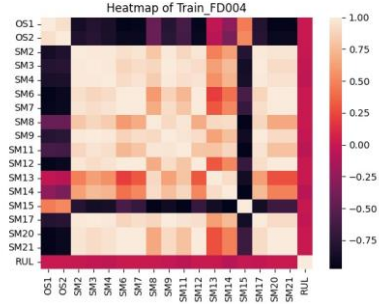

(d)

**Fig. 1**: Heat map of (a)FD001, (b)FD002, (c)FD003, (d)FD004 train datasets

According to Figure 1 of FD001, the heat-map shows the strongest relationships consisting of SM2, SM3, SM4, SM7, SM8, SM11, SM12, SM13, SM15, SM17, SM20, SM21 are correlated with almost all sensor measurement except SM6, SM9, and SM14. Regarding the trend, it seems like RUL has a similar correlation trend with SM7, SM12, SM20, and SM21, which can be implied that RUL is mainly determined by SM7, SM12, SM20, and SM21. The other interesting part is that OS1 and OS2 do not have any relationship with the sensor measurements. In FD002, the RUL has no relationship with any of the sensor measurements. SM15 has the strongest relationship with every other sensor measurement, but it seems like it has the opposite trend compared to others. Every operational setting and sensor measurements are correlated with each other except for the SM13 and SM14. Unlike FD001, FD002 shows OS1 and OS2 are affected by most of the sensor measurements. In FD003, the heatmaps show similar trends and relationships among the operational setting and sensor measurements as FD001. For FD004, the heatmaps show similar trends and relationships among the operational setting and sensor measurements as FD002.

### 3.2 Data Preparation

The original data contains three operational settings and sensor measurements from column variables 6 to 26, with unclear descriptions. The operational settings are Mach Number (0 to 0.90), altitude (sea level to 40,000 feet), and sealevel temperature (−60F to 103F). Upon further research (Saxena et al., 2008), the labels were identified for the sensor measurements with an explanation of each label provided in Appendix A.

From the original C-MAPSS datasets, training sets consist of all data up until each failure and testing sets have the cutoff of some data before the failure. For example, in the train set for FD001, all rows with unit = 1 are the data from the first failure. The last row of unit 1 has cycles = 192 which means the engine failed after 192 operational cycles. For the testing set, the last row of unit = 1 has cycle = 31, which is not when the failure happened. The separate RUL files contain the number of cycles until failure from the last sample for each test unit. In the case of the test set for FD001 unit 1, it is cycles = 112.

## 4 Methodology

### 4.1 Federated Learning and FedAvg Aggregation

Engine fault prediction will be performed using an FL approach (McMahan et al., 2017). FL is a specialized form of distributed learning that places a strong emphasis on data privacy and security. It is particularly beneficial in scenarios where data confidentiality is crucial in a highly sensitive environment. This approach is chosen because it helps prevent data leakage and reverse engineering of the data. However, the setup can lead to biased subsets of data at each node, as the training data is not shareable. To address this, a federated learning algorithm is employed.

Initially, the server sends instructions to each node to train a local model. These local node models train on their respective data and after a training round, they only transmit their updated weights to the central server. The central server then aggregates these weights. The most common linear aggregation method is Federated Averaging (FedAvg). FedAvg is a generalized version of local-SGD (Wang et al., 2021), which performs a weighted average of local model parameters (weights) after a certain number of optimization steps are performed by each model. The weights for the average are determined as seen fit by the implementation. In our case, we chose to start with a simple method, where the weight is the number of data samples available for training at each node. The combined model is then transmitted back to the nodes, which use the updated model parameters as a starting point for another round of training (Qi et al., 2024). This process continues for multiple rounds until the global model converges.

## 4.2 Long Short-Term Memory (LSTM)

Upon reviewing the current work being done for PM, especially the work done in (Asif et al., 2022), we decided that the best technique to model RUL is to use deep learning with LSTM neural network. LSTM is a type of Recurrent Neural Network (RNN), which are suitable for modeling events that happen in sequence. LSTM avoids the common vanishing gradient problem (Hochreiter, 1998) by using gates to store or forget information as needed. This allows LSTM to learn longterm dependencies more effectively while having the ability to forget learned relationships that no longer benefit the goal of minimizing loss. Each LSTM cell is composed of three gates implemented as sigmoid functions. Data $x_t$ comes into the cell and becomes part of $g_t$, the candidate cell which also has previous memory information. The $f_t$ (forget) gate decides what information should be forgotten, the $i_t$ (input) gate decides what information should be stored and the $o_t$ (output) gate decides what information should be the output from the cell. Hidden state $S_{t-1}$ and cell state $L_{t-1}$ contain the short and long-term memory states from the previous cells respectively. Figure 2 demonstrates the internal workings of a single LSTM cell.

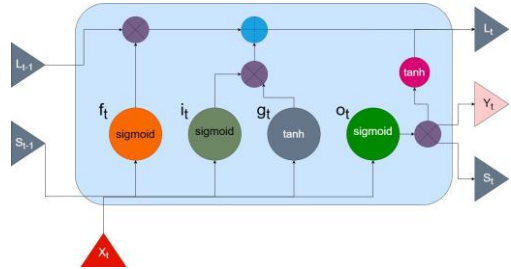

**Fig. 2**: Diagram of LSTM cell internal structure.

## 4.3 Model Validation

We used the KerasTuner python module to do extensive tuning of our neural networks. This module allows nearly limitless customization of model tuning with skilled use of Python. See the documentation of the KerasTuner Python module for more information on the usage. The table below shows the exact hyperparameters and values tested.

**Table 2**: Keras Tuner Configuration

| Hyper-parameter Name | Values |
| --- | --- |
| Sequence Length | 1, 2, 4, 8 |
| Batch Size | 1, 8, 32 |
| Layer Dropout | 0.1, 0.2, 0.3 |
| Recurrent Dropout | 0.1, 0.2 |
| Learning Rate | 0.0001, 0.001, 0.002 |
| Gaussian Noise | 0.01, 0.1 |

### 4.4 Metrics and Loss

When it comes to metrics for regression problems in machine learning, there are three main loss functions: Mean Squared Error (MSE), Mean Absolute Error (MAE) and R-squared ($R^2$). Let $N$ be the number of prediction samples, $\bar{y}$ be the mean of the true values, $y_i$ and $\hat{y}_i$ be the true and predicted values at sample $i$ respectively. These metrics are defined as:

$$MSE = \frac{\sum_{i=1}^{N}(\hat{y}_i - y_i)^2}{N} \quad (1)$$

$$MAE = \frac{\sum_{i=1}^{N}|\hat{y}_i - y_i|}{N} \quad (2)$$

$$R^2 = 1 - \frac{\sum_{i=1}^{N}(\hat{y}_i - y_i)^2}{\sum_{i=1}^{N}(\bar{y} - y_i)^2} \quad (3)$$

MSE is a very punishing loss because it is non-linear with respect to error size. For example, an error of 10 is 100 times worse than an error of 1 because of the squaring. Additionally, the use of the square changes the units of the error to squared target units which may be difficult to understand. Because of this, Root-MSE is a popular choice:

$$RMSE = \sqrt{MSE}$$

MAE can be good because it is recorded in the same units as the target variable (engine cycles in our case), however, it is linearly punishing and treats an error of $q\beta$ as being $q$ times worse than an error of $\beta$ which may not be good in certain cases such as jet engine failure. $R^2$ is generally not chosen as a loss as it is difficult to understand it as a loss function, and easy to manipulate by adding more variables to our input and "curating" the validation data to have low variance. However, given two models with the same input variables evaluated on the same data, $R^2$ can be a good "tiebreaker" for similar RMSE or MAE. For the reasons laid out above, we selected to use RMSE for the training and validation loss of our models, but we reported MAE as our second metric.

### 4.5 Model Architecture

Derived from the work done in (Asif et al., 2022), we used a neural network architecture that leverages the recurrent capabilities and robustness of LSTM cells with the approximation abilities of fully connected layers. The dropout layers and recurrent dropout in the LSTM are used to prevent over-fitting and improve generalization on unseen data. The Gaussian noise layers are also generalization-aiding layers, as well as the differential privacy facet of Federated Learning.

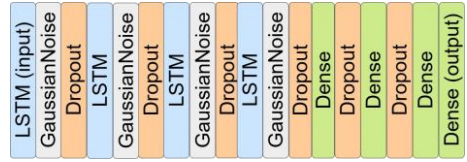

**Fig. 3**: Neural Network Architecture Diagram.

## 4.6 Preprocessing

Preprocessing data effectively is crucial for ensuring optimal performance in machine learning models, and it involves three key techniques: filtering, pruning units, and scaling.

Many of the signals that make up the input of this dataset have noise included. The noise was purposefully added when the data was being generated (Saxena et al., 2008). Noise can be bad for machine learning models, since the models may over-fit the noise of the training data, which has no real influence on the underlying phenomena. Filtering is a way to reduce the noise in the signals and improve the model generalization. We chose to use a median filter (Kaur et al., 2019; Kumar and Sodhi, 2020), which is a sliding window filter that suppresses noise and outliers.

To select the kernel size for filtering each signal, we divided the signals based on each unit within each node. Then we tried kernel sizes $k$ = 3,5,7,9,11,...,$M_k$ where $M_k = \lfloor \frac{s_i}{10} \rfloor$, $s_i$ being the number of samples for unit $i$. The best kernel for each unit was selected as the one that maximizes the linear correlation between the target (RUL) and the signal being analyzed. We then took the weighted geometric average of all the kernels and set it as the final kernel size to use. The weight varied as the correlation and the number of samples for the unit. The best was selected depending on the final model performance.

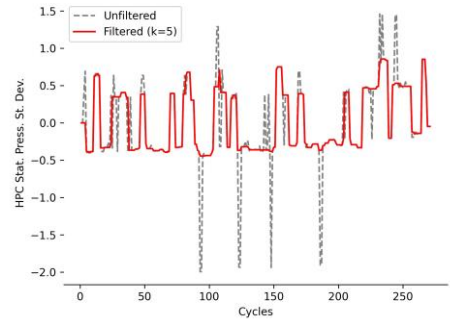

**Fig. 4**: Example of Median Filter on HPC Static Pressure Signal for FD002 Engine 11 Time series.

The second technique used is a simple "pruning" technique, which randomly removed some samples from the ends of the sequences of some of the time series in the training set. This method is similar in execution to neural network pruning (Li et al., 2019; Pasandi et al., 2020) though, the goal is to make the data more representative as opposed to reducing complexity. This was done to mimic the test set time series, which did not run until failure. The general idea was to prevent the model from over-fitting to time series that start at full health and end at critical failure. Listing 1, is the code that was used for pruning.

Scaling inputs is important when dealing with data in vastly different scales, such as data taken from different sensors from different jet engines in different operating conditions (Sharma, 2022). The differing scales and variances can cause the neural network to unfairly bias certain dimensions in the data due to undesirable reasons. We elected to use the popular z-score normalization technique, which takes all dimensions in the data and normalizes them to have approximately zero mean and standard deviation of 1. This creates an even playing field when it comes to assigning neural network parameters during optimization. To use this technique, we simply turn each observation in each sample into a z-score. Let $X$ be the set of all observations for a certain feature in one of our agents, we can find the z-score for the $i$-th observation $x_i$ using:

$$z_i = \frac{x_i - \bar{X}}{\sigma_X}$$

Overall, these pre-processing steps help ensure the data is free from noise, outliers, and scale-related biases.

**Listing 1** Pruning Units
```python
import numpy as np
import pandas

def prune_units(df, key="unit", prune_chance=0.3, p=0.4, pplus=0.1):
    for _, data in df.groupby(key):
        chance = np.random.rand()
        rows = data.shape[0]
        if chance <= pplus:
            prune_rows = rows*0.75
            df.drop( data.tail(int( prune_rows
                            )).index,
                    inplace=True)
        elif chance <= prune_chance:
            prune_rows = rows*p
            df.drop( data.tail(int(
                    prune_rows
                            )).index,
                    inplace=True)
```

## 4.7 Feature Engineering

Given that our dataset is made up of multiple time series that deal with the degradation of parts, we decided to use some feature engineering techniques to add dimensions and new predictors to our data. The first technique is a simple accumulation technique where we take the cumulative sum of certain signals throughout the degradation process. The second technique is explained as follows.

As proposed in (Mu et al., 2016), taking the derivative of the input space can add helpful dimensions to our dataset. We took a similar approach by taking a simple, single-dimension, single-time-step signed change calculation to some of our features and added these as new features. The single time step is because we do not know how long the "cycle" time dimension is. Given that most engines in the dataset fail within 200 cycles it is safe to say each cycle is a relatively large time step such as hours, or even an entire flight. Thus, having too large of a window may not capture meaningful change. The following snippet of code demonstrates how the rate of change is calculated for one dimension and time-step = $dt$:

**Listing 2** Rate of Change
```python
import numpy as np
import pandas as pd

def derive(data, feature, dt):
    x = data[feature].to_numpy()
    dx = data[feature].to_numpy()
    # first dt-1 samples assumed
    # to have no change in x

    return dx
```

```
    dx[:dt] = 0 for i in range(dt,data.shape[0]):
    dx[i] = (x[i] - x[i-dt])/dt
```

# 5 Results and Discussion

## 5.1 Model Hyperparameters

The following are the final model parameters based on Keras Tuner experimentation:

**Table 3**: Hyperparameter Selection

| Hyper-parameter Name | Best Value |
|---|---|
| LSTM Layers | 4 |
| Dense Layers | 4 |
| Units Per Layer | 64 |
| Batch Size | 32 |
| Layer Dropout | 0.1 |
| Recurrent Dropout | 0.2 |
| Learning Rate | 0.001 |
| Gaussian Noise | 0.01 |

## 5.2 Model Performance

Below is a table detailing the validation and testing set performance for our agents and our aggregated model. We also have some graphical demonstrations of how our model predicts the degradation of the different faults over time, considering the different conditions simulated within the different agents.

**Table 4**: Validation and Test Performance

| Agent | Validation RMSE | Test RMSE |
|---|---|---|
| **FD001** | 13.7014 | 13.811 |
| **FD002** | 14.6618 | 20.587 |
| **FD003** | 14.3186 | 14.201 |
| **FD004** | 18.1955 | 22.998 |
| **Aggregated** | 15.2193 | 17.899 |

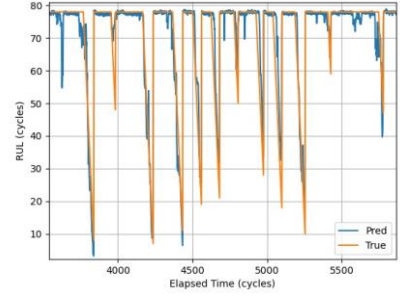

(a)

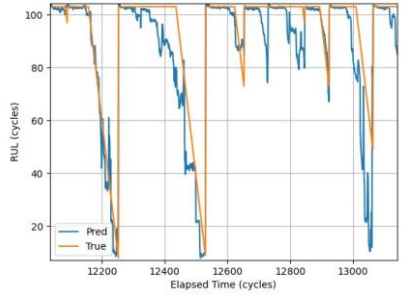

(b)

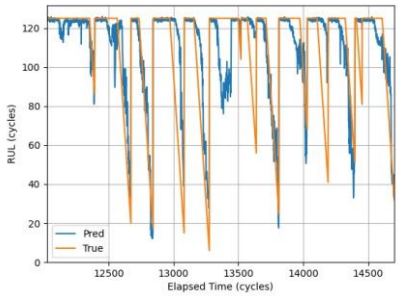

(c)

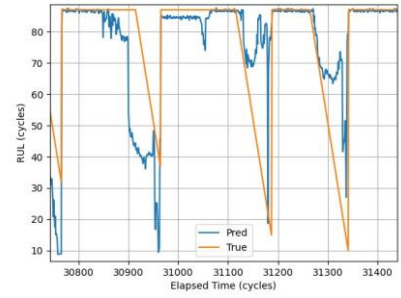

(d)

**Fig. 5**:

## 5.3 Statistical Analysis

We averaged the results across the different agents over several studies aggregated by Asif *et. al.*, and compared the results with our federated averaged model results. We used statistical tests to compare these models. First, we ran twotailed, equal means tests for all of them, without selecting an $\alpha$ value beforehand. All of the p-values were extremely low, thus all of the equal means hypotheses were rejected, suggesting that our models either outperformed or under-perform against each study.

We then performed greater-than-orequal-to, and less-than-or-equal-to onetailed tests against each of the studies and recorded the results. The table below summarizes the studies for which our model was statistically better on average. The study number refers to the order in which the study appears in Table 9 of the paper by Asif *et. al.*

**Table 5**: Statistical Model Comparison

| Study # | Null Hyp. ($H_0$) | t statistic | Reject $H_0$? |
|---|---|---|---|
| 13 | $\mu$<= 18.44 | -14.35 | No |
| 3 | $\mu$<= 18.86 | -23.70 | No |
| 5 | $\mu$<= 19.24 | -32.37 | No |
| 10 | $\mu$<= 19.29 | -33.39 | No |
| 2 | $\mu$<= 21.24 | -77.50 | No |

*Note: all p-values 0.99999999 or higher.*

We also generated 95% confidence intervals for the true mean error of our agent and aggregated models.

## 5.4 Discussion

The experiment results validated our hypothesis, as on average the FL approach either maintained or improved the model performance. While the test error across the individual agents showed significant variance, the aggregated model demonstrated the ability to generalize and balance the error overall. The variance in individual agent performance can explained by the operating conditions detailed in Table 1. FD002 and FD004 are simulated with variance in six operating conditions (altitude, pressure, airspeed, etc.) while FD001 and FD003 are simulated at sea level with no variance. Thus, FD002 and FD004 performed 1.25-1.6 times worse than the other agents. The statistical analysis provided further insights into the effectiveness of the FL model. By conducting two-tailed equal means tests and greater-than-orequal-to and less-than-or-equal-to onetailed tests, we were able to compare our model's performance against other studies. The consistently low p-values and the rejection of null hypotheses in several cases suggest that our FL-based model either outperforms or is comparable to existing methods in predictive maintenance. This statistical rigor confirms that the federated approach not only preserves data privacy but also maintains, if not improves, model accuracy.

**Table 6**: Confidence Intervals for Mean Model Errors

| Agent Name | Lower Bound | Upper Bound |
|---|---|---|
| | Testing set predictions for (a)FD001, (b)FD002, (c)FD003, (d)FD004 | |
| FD001 | 13.10 | 13.44 |
| FD002 | 20.55 | 21.16 |
| FD003 | 13.82 | 13.98 |
| FD004 | 22.90 | 23.48 |
| Aggregated | 17.71 | 17.90 |

## 6 Conclusion

This study explores the application of FL in PM tasks, focusing on engine fault prediction using deep learning with LSTM cells. The primary motivation behind this work was to address the challenges of maintaining data privacy while allowing collaboration between entities with data silos. Our main hypothesis was that implementing the FL framework would either maintain or improve the model performance on average when compared to the centralized learning approach. The implementation of FL offers several key advantages in the context of predictive maintenance. Firstly, it preserves data privacy by ensuring that sensitive data remains localized on the devices where it is generated. This is particularly important in industries like aerospace, where operational data can include proprietary and sensitive information. By minimizing the need for data centralization, FL reduces the risk of data breaches while allowing collaboration between entities that would otherwise not be able to collaborate. However, the study also highlights some challenges associated with the implementation of FL in predictive maintenance. Managing the FL process across multiple devices introduces complexity, particularly in coordinating updates and maintaining consistent model performance across diverse environments. Additionally, the federated approach can lead to biased subsets of data at each node, which may affect the generalization of the global model. Despite these challenges, the ability to collaborate across different entities without sharing raw data opens up new possibilities for improving maintenance models industry-wide. This collaborative approach can accelerate advancements in predictive maintenance technologies, benefiting not just individual companies but the entire industry.

## 7  Declarations

- This project has no external funding
- All authors consent for publication
- Data acquired from NASA PCoE under Public Domain License
- Code available upon request

# Appendix A  Column Labels and Units

Table A1: Sensor Descriptions

| Original Label | New Label | Sensor Name | Units | Descriptions |
|---|---|---|---|---|
| SM1 | fan in temp | T2 | R | Total temp at fan inlet |
| SM2 | lpc out temp | T24 | R | Total temp at LPC outlet |
| SM3 | hpc out temp | T30 | R | Total temp at HPC outlet |
| SM4 | lpt out temp | T50 | R | Total temp at LPT outlet |
| SM5 | fan in press | P2 | psia | Pressure at fan inlet |
| SM6 | bypass press | P15 | psia | Total pressure in bypass duct |
| SM7 | hpc out press | P30 | psia | Total pressure at HPC outlet |
| SM8 | fan speed | Nf | rpm | Physical fan speed |
| SM9 | core speed | Nc | rpm | Physical core speed |
| SM10 | epr | epr | N/A | Engine pressure ratio (P50/P2) |
| SM11 | hpc stat press | Ps30 | psia | Static pressure at HPC outlet |
| SM12 | flow press ratio | Phi | pps/psi | Ratio of fuel flow to Ps30 |
| SM13 | corr fan speed | NRf | rpm | Corrected fan speed |
| SM14 | corr core speed | NRc | rpm | Corrected core speed |
| SM15 | bypass ratio | BPR | N/A | Bypass ratio |
| SM16 | burner fuel ratio | farB | N/A | Burner fuel-air ratio |
| SM17 | bleed enthalpy | htBleed | N/A | Bleed enthalpy |
| SM18 | dmd fan speed | Nf dmd | rpm | Demanded fan speed |
| SM19 | dmd corr fan speed | PCNfR dmd | rpm | Demanded corrected fan speed |
| SM20 | hpt bleed | WC31 | lbm/s | HPT coolant bleed |
| SM21 | lpt bleed | WC32 | lbm/s | LPT coolant bleed |